\documentclass{llncs}

\usepackage{array}
\usepackage{moreverb}
\usepackage{tabularx}
\usepackage{graphicx}
\usepackage{amsmath}
\usepackage{hyperref}

\graphicspath{{figures/}}

\def\RR{{\rm I\hspace{-0.50ex}R}}

\begin{document}

\mainmatter

\title{Genetic Programming for Kernel-based Learning with Co-evolving Subsets Selection}
\titlerunning{GP for Kernel-based Learning}

\author{Christian Gagn\'e\inst{1,2} \and Marc Schoenauer\inst{2} \and\\
Mich\`ele Sebag\inst{2} \and Marco Tomassini\inst{1}}
\authorrunning{Gagn\'e {\em et al.}}
\tocauthor{Christian Gagn\'e (Universit\'e de Lausanne/INRIA), Marc Schoenauer (INRIA), Mich\`ele Sebag (CNRS), Marco Tomassini (Universit\'e de Lausanne)}
\institute{
Information Systems Institute,\\
Universit\'e de Lausanne, CH-1015 Dorigny, Switzerland.\\
\email{\{christian.gagne,marco.tomassini\}@unil.ch}\\[1.2em]
\and
\'Equipe TAO -- INRIA Futurs / CNRS UMR 8623,\\
LRI, Bat. 490, Universit\'e Paris Sud, 91405 Orsay CEDEX, France.\\
\email{\{marc.schoenauer,michele.sebag\}@lri.fr}
}

\maketitle

\begin{abstract}
Support Vector Machines (SVMs) are well-established Machine Learning (ML) algorithms. They rely on the fact that i) linear learning can be formalized as a well-posed optimization problem; ii) non-linear learning can be brought into linear learning thanks to the {\em kernel trick} and the mapping of the initial search space onto a high dimensional feature space. The kernel is designed by the ML expert and it governs the efficiency of the SVM approach. In this paper, a new approach for the automatic design of kernels by Genetic Programming, called the Evolutionary Kernel Machine (EKM), is presented. EKM combines a well-founded fitness function inspired from the margin criterion, and a co-evolution framework ensuring the computational scalability of the approach. Empirical validation on standard ML benchmark demonstrates that EKM is competitive using state-of-the-art SVMs with tuned hyper-parameters.
\end{abstract}

\section{Introduction}
\label{sec:Introduction}

Kernel methods, including the so-called Support Vector Machines (SVMs), are well-established learning approaches with both strong theoretical foundations and successful practical applications \cite{Shawe2004}. SVMs rely on two main advances in statistical learning. First, the linear supervised machine learning task is set as a well-posed (quadratic) optimization problem. Second, the above setting is extended to non-linear learning via the {\em kernel trick}\,: given a (manually designed) change of representation $\Phi$ mapping the initial space onto the so-called feature space, linear hypotheses are characterized in terms of the scalar product in the feature space, or {\em kernel}. These hypotheses correspond to non-linear hypotheses in the initial space. Although many specific kernels have been proposed in the literature, designing a kernel well suited for an application domain or a dataset so far remains an art more than a science.

This paper proposes a system, the {\em Evolutionary Kernel Machine} (EKM), for the automatic design of data-specific kernels. EKM applies Genetic Programming (GP) \cite{Koza1992} to construct symmetric functions (kernels), and optimizes a fitness function inspired from the margin criterion \cite{Gilad-Bachrach2004}. Kernels are assessed within a Nearest Neighbor classification process \cite{Duda2001,Yu2002}. In order to cope with computational complexity, a cooperative co-evolution governs the prototype subset selection and the GP kernel design, while the fitness case subset selection undergoes a competitive co-evolution.

The paper is organized as follows. Section \ref{sec:KernelBasedLearning} introduces the formal background and notations on kernel methods. Sections \ref{sec:GPKernel} and \ref{sec:Fitness} respectively describe the GP representation and the fitness function proposed for the EKM. Scalability issues are addressed in the co-evolutionary framework introduced in Section \ref{sec:CSS}. Results on benchmark problems are given in Section \ref{sec:Results}. Finally, related works are discussed in Section \ref{sec:RelatedWorks} before concluding the paper in Section \ref{sec:Conclusion}.

\section{Formal Background and Notations}
\label{sec:KernelBasedLearning}

Supervised machine learning takes as input a dataset $\mathcal{E} = \{(\mathbf{x}_i,y_i),~i=1\ldots n$, $\mathbf{x}_i\in X,~y_i\in Y\}$, made of $n$ examples; $\mathbf{x}_i$ and $y_i$ respectively stand for the description and the label of the $i$-th example. The goal is to construct a hypothesis $\mathrm{h}(\mathbf{x})$ mapping $X$ onto $Y$ with minimal generalization error. Only vectorial domains ($X = \RR^d$) are considered throughout this paper; further, only binary classification problems ($Y=\{1,-1\}$) are considered in the rest of this section. 

Due to space limitations, the reader is referred to \cite{Cristianini2000} for a comprehensive presentation of SVMs. In the simplest (linear separable) case, the hyper-plane $\mathrm{h}(\mathbf{x})$ maximizing the geometrical margin (distance to the closest examples) is constructed. The label associated to example $x$ is the sign of $\mathrm{h}(\mathbf{x})$, with:
\[ \mathrm{h}(\mathbf{x}) = \sum_i \alpha_i <\mathbf{x},\mathbf{x}_i> +\,b \]
where $<\mathbf{x},\mathbf{x}_i>$ denotes the scalar product of $\mathbf{x}$ and $\mathbf{x}_i$. Let $\Phi$ denotes a mapping from the instance space $X$ onto the feature space and let the kernel $\mathrm{K}(\mathbf{x},\mathbf{x}')$ be defined as:
\[ \mathrm{K}: X\times X\mapsto\RR;\hspace*{1in} \mathrm{K}(\mathbf{x},\mathbf{x}') = <\Phi(\mathbf{x}),\Phi(\mathbf{x}')> \]
Under some conditions (the kernel trick), non-linear classifiers on $X$ are constructed as in the linear case, and characterized as $\mathrm{h}(\mathbf{x}) = \sum_i \alpha_i \mathrm{K}(\mathbf{x},\mathbf{x}_i) + b$.

Besides SVMs, the kernel trick can be used to revisit all learning methods involving a distance measure. In the paper, the kernel nearest neighbor (Kernel-NN) algorithm \cite{Yu2002}, which revisits the $k$-nearest neighbors ($k$-NN) \cite{Duda2001}, is considered. Given a distance (or dissimilarity) function $\mathrm{d}(\mathbf{x},\mathbf{x}')$ defined on the instance space $X$, given a set of labelled examples $\mathcal{E}=\{(\mathbf{x}_1,y_1),\ldots,(\mathbf{x}_n,y_n)\}$ and an instance $\mathbf{x}$ to be classified, the $k$-NN algorithm: i) determines the $k$ examples closest to $\mathbf{x}$ according to $\mathrm{d}(\mathbf{x},\mathbf{x}')$; ii) outputs the majority class of these $k$ examples. Kernel-NN proceeds as $k$-NN, where distance $\mathrm{d_K}(\mathbf{x},\mathbf{x}')$ is defined after the kernel $\mathrm{K}(\mathbf{x},\mathbf{x}')$ (more on this in Section \ref{sec:Fitness}). 

Standard kernels on $X=\RR^d$ include Gaussian and polynomial kernels\footnote{Respectively $\mathrm{K}(\mathbf{x},\mathbf{x}')=\exp{\left(-\frac{\|\mathbf{x}-\mathbf{x}'\|^2}{\sigma}\right)}$ and $\mathrm{K}(\mathbf{x},\mathbf{x}')=(<\mathbf{x},\mathbf{x}'>+\,c)^k$}. It must be noted that the addition, multiplication and compositions of kernels are kernels, and therefore the standard SVM machinery can find the optimal value of hyper-parameters (e.g. $\sigma, c$ or $k$) among a finite set. Quite the opposite, the functional (symbolic) optimization of $\mathrm{K}(\mathbf{x},\mathbf{x}')$ cannot be tackled to our best knowledge except by Genetic Programming.

\section{Genetic Programming of Kernels}
\label{sec:GPKernel} 

The Evolutionary Kernel Machine applies GP to determine symmetric functions $\mathrm{K}(\mathbf{x},\mathbf{x}')$ on $\RR^d\times\RR^d$ best suited to the dataset at hand. As shown in Table \ref{tab:GPPrimitives}, the main difference compared to standard symbolic regression is that terminals are symmetric expressions of $\mathbf{x}$ and $\mathbf{x'}$ (e.g. $x_i + x'_i$, or $x_i x_j' + x_j x_i'$), enforcing the symmetry of the kernels ($\mathrm{K}(\mathbf{x},\mathbf{x'})~=~\mathrm{K}(\mathbf{x'},\mathbf{x})$).

The initialization of GP individuals is done using a ramped half and half procedure \cite{Koza1992}. The selection probability of terminals $A_i, M_i$, $I_i$ and $S_i$ (respectively $C_{i,j}$) is divided by $1/d$ (resp. $2/d(d+1)$), where $d$ is the dimension of the initial instance space ($X=\RR^d$).
\begin{table}[tb]
\caption{GP primitives involved in the kernel functions $\mathrm{K}(\mathbf{x},\mathbf{x'}),~\mathbf{x},\mathbf{x'}\in\RR^d$.}
\label{tab:GPPrimitives}
\begin{center}
\begin{tabularx}{\linewidth}{c|c|X}
Name & \# args. & Description \\\hline
ADD2 & $2$      & Addition of two values, $\mathrm{f_{ADD2}}(a_1,a_2)=a_1+a_2$.\\
ADD3 & $3$      & Addition of three values, $\mathrm{f_{ADD3}}(a_1,a_2,a_3)=a_1+a_2+a_3$.\\ 
ADD4 & $4$      & Addition of four values, $\mathrm{f_{ADD4}}(a_1,a_2,a_3,a_4)=a_1+a_2+a_3+a_4$.\\ 
SUB  & $2$      & Subtraction, $\mathrm{f_{SUB}}(a_1,a_2)=a_1-a_2$.\\
MUL2 & $2$      & Multiplication of two values, $\mathrm{f_{MUL2}}(a_1,a_2)=a_1 a_2$.\\
MUL3 & $3$      & Multiplication of three values, $\mathrm{f_{MUL3}}(a_1,a_2,a_3)=a_1 a_2 a_3$.\\
MUL4 & $4$      & Multiplication of four values, $\mathrm{f_{MUL4}}(a_1,a_2,a_3,a_4)=a_1 a_2 a_3 a_4$.\\
DIV  & $2$      & Protected division, $\mathrm{f_{DIV}}(a_1,a_2)=\left\{\begin{array}{c@{\qquad}c}1 & |a_2|<0.001\\ a_1/a_2 & \mbox{otherwise}\end{array}\right.$.\\
MAX  & $2$      & Maximum value, $\mathrm{f_{MAX}}(a_1,a_2)=\max(a_1,a_2)$.\\
MIN  & $2$      & Minimum value, $\mathrm{f_{MIN}}(a_1,a_2)=\min(a_1,a_2)$.\\
EXP  & $1$      & Exponential value, $\mathrm{f_{EXP}}(a)=\exp(a)$.\\
POW2 & $1$      & Square power, $\mathrm{f_{POW2}}(a)=a^2$.\\\hline
$A_i,~i=1\ldots d$ & $0$ & Add the $i^{th}$ components, $x_i+x'_i$.\\
$M_i,~i=1\ldots d$ & $0$ & Multiply the $i^{th}$ components, $x_i x'_i$.\\
$S_i,~i=1\ldots d$ & $0$ & Maximum between the $i^{th}$ components, $\max(x_i,x'_i)$.\\
$I_i,~i=1\ldots d$ & $0$ & Minimum between the $i^{th}$ components, $\min(x_i,x'_i)$.\\
$C_{i,j},~\begin{array}[t]{c}i=1\ldots d\\ j=1\ldots i\end{array}$ & $0$ & Crossed multiplication-addition between the $i^{th}$ and $j^{th}$ components, $(x_i x'_j + x_j x'_i)$.\\
DOT  & $0$      & Scalar product of $\mathbf{x}$ and $\mathbf{x'}$, $<\mathbf{x},\mathbf{x'}>$.\\
EUC  & $0$      & Euclidean distance of $\mathbf{x}$ and $\mathbf{x'}$, $\|\mathbf{x}-\mathbf{x'}\|$.\\
E    & $0$      & Ephemeral random constants, generated uniformly in $[-1,1]$.\\
\end{tabularx}
\end{center}
\end{table}

Indeed the kernel functions built after Table \ref{tab:GPPrimitives} might not satisfy Mercer's condition ($\mathrm{K}(\mathbf{x},\mathbf{x})\le 0 \not \Rightarrow \mathbf{x}=0$) required for SVM optimization \cite{Cristianini2000}. However these kernels will be assessed along a Kernel-NN classification rule \cite{Yu2002}; therefore the fact that they are not necessarily positive is not a limitation. Quite the contrary, EKM kernels can achieve feature selection; typically, terminals associated to non-informative features should disappear along evolution. The use of EKM for feature selection will be examined in a future work.

\section{Fitness Measure}
\label{sec:Fitness}

Every kernel $\mathrm{K}(\mathbf{x},\mathbf{x}')$ is assessed after the Kernel-NN classification rule, using the dissimilarity $\mathrm{d_{K}}$ defined as
\[ \mathrm{d_{K}}(\mathbf{x},\mathbf{x}')^2 = \mathrm{K}(\mathbf{x},\mathbf{x})+\mathrm{K}(\mathbf{x}',\mathbf{x}')-2\mathrm{K}(\mathbf{x},\mathbf{x}') \]
  
Given a prototype set $\mathcal{E}_p=\{(\mathbf{x}_1,y_1),\ldots,(\mathbf{x}_\ell,y_\ell)\}$ and a training example $e=(\mathbf{x},y)$, let us assuming that $\mathcal{E}_p$ is ordered by increasing dissimilarity to $\mathbf{x}$ ($\mathrm{d_{K}}(\mathbf{x},\mathbf{x}_i)\le \mathrm{d_{K}}(\mathbf{x},\mathbf{x}_{i+1})$). Let $\mathrm{p}(e)$ denotes the minimum rank over all prototype examples in the same class as $e$ ($\mathrm{p}(e)=\min\{i,~y_i=y,~i=1\ldots\ell\}$); let $\mathrm{n}(e)$ denotes the minimum rank over all other prototype examples (not belonging to the same class as $e$, $\mathrm{n}(e)=\min\{i,~y_i\not =y,~i=1\ldots\ell\}$).

As noted by \cite{Gilad-Bachrach2004}, the quality of the Kernel-NN classification of $e$ can be assessed from $\delta_\mathrm{K}(e)=\mathrm{n}(e)-\mathrm{p}(e)$. The higher $\delta_\mathrm{K}(e)$, the more confident the classification of $e$ is, e.g. with respect to perturbations of $\mathcal{E}_p$ or $\mathrm{d_{K}}$; $\delta_\mathrm{K}(e)$ measures the margin of $e$ with respect to Kernel-NN. 

Accordingly, given a prototype set $\mathcal{E}_p=\{(\mathbf{x}_1,y_1),\ldots,(\mathbf{x}_\ell,y_\ell)\}$ and a fitness case subset $\mathcal{E}_s=\{(\mathbf{x}'_1,y'_1),\ldots,(\mathbf{x}'_m,y'_m)\}$, the fitness function associated to $\mathrm{K}(\mathbf{x},\mathbf{x}')$ is defined as
\[ \mathrm{F}(\mathrm{K}) = \frac{1}{m} \sum_{i=1}^m \delta_\mathrm{K}(\mathbf{x}'_i,y'_i) - \ell\]

The computation of $\mathrm{F}$ has linear complexity in the number $\ell$ of prototypes and in the number $m$ of fitness cases.  In a standard setting, $\mathcal{E}_p$ and $\mathcal{E}_s$  both coincide with the whole training set $\mathcal{E}$ ($\ell=m=n$). However the quadratic complexity of the fitness computation with respect to the number $n$ of training examples is incompatible with the scalability of the approach.

\section{Tractability Through Co-evolution}
\label{sec:CSS}

EKM scalability is obtained along two directions, by i) reducing the number $\ell$ of prototypes used for classification, and ii) reducing the size $m$ of the fitness case subset considered during each generation.

\begin{figure}[tb]
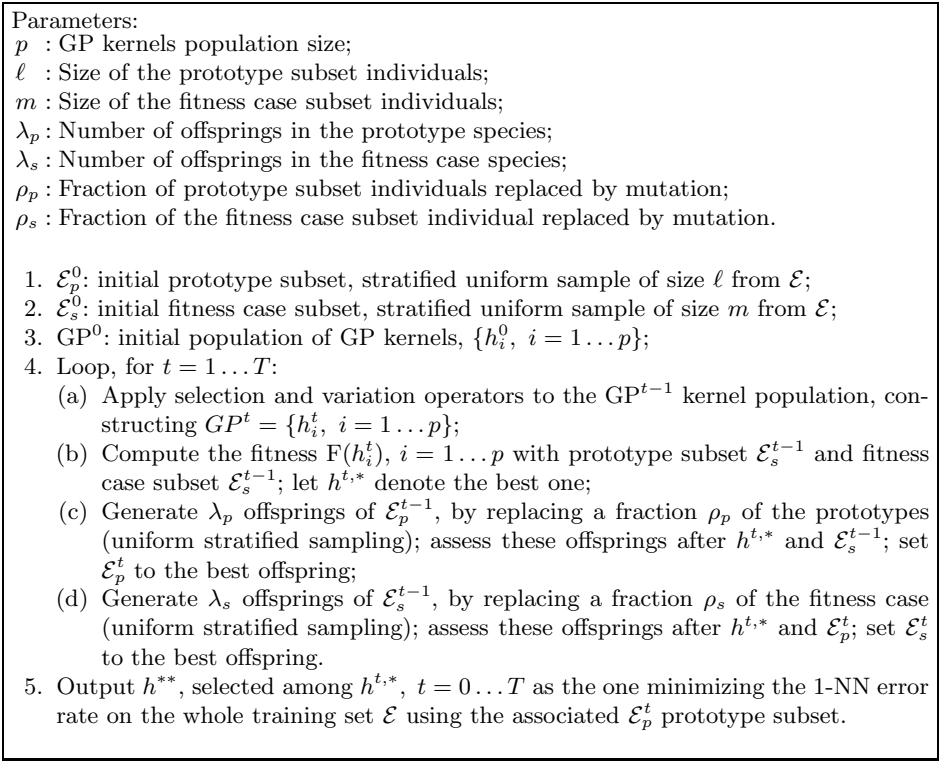

\begin{center}\fbox{\parbox{\linewidth}{
\small
Parameters:\\
\begin{tabular}{l@{\,:\;}l}
$p$         & GP kernels population size;\\
$\ell$      & Size of the prototype subset individuals;\\
$m$         & Size of the fitness case subset individuals;\\
$\lambda_p$ & Number of offsprings in the prototype species;\\
$\lambda_s$ & Number of offsprings in the fitness case species;\\
$\rho_p$    & Fraction of prototype subset individuals replaced by mutation;\\
$\rho_s$    & Fraction of the fitness case subset individual replaced by mutation.\\
\end{tabular}\\
\begin{enumerate}
\item $\mathcal{E}_p^0$: initial prototype subset, stratified uniform sample of size $\ell$ from $\mathcal{E}$;
\item $\mathcal{E}_s^0$: initial fitness case subset, stratified uniform sample of size $m$ from $\mathcal{E}$;
\item GP$^0$: initial population of GP kernels, $\{h^0_i,~i=1\ldots p\}$;
\item Loop, for $t=1\ldots T$:
\begin{enumerate}
\item Apply selection and variation operators to the GP$^{t-1}$ kernel population, constructing $GP^t=\{h^t_i,~i=1\ldots p\}$;
\item Compute the fitness $\mathrm{F}(h^t_i)$, $i=1\ldots p$ with prototype subset $\mathcal{E}_s^{t-1}$ and fitness case subset $\mathcal{E}_s^{t-1}$; let $h^{t,*}$ denote the best one;
\item Generate $\lambda_p$ offsprings of $\mathcal{E}_p^{t-1}$, by replacing a fraction $\rho_p$ of the prototypes (uniform stratified sampling); assess these offsprings after $h^{t,*}$ and $\mathcal{E}_s^{t-1}$; set $\mathcal{E}_p^t$ to the best offspring;
\item Generate $\lambda_s$ offsprings of $\mathcal{E}_s^{t-1}$, by replacing a fraction $\rho_s$ of the fitness case (uniform stratified sampling); assess these offsprings after $h^{t,*}$ and $\mathcal{E}_p^t$; set $\mathcal{E}_s^{t}$ to the best offspring.
\end{enumerate}
\item Output $h^{**}$, selected among $h^{t,*},~t =0\ldots T$ as the one minimizing the 1-NN error rate on the whole training set $\mathcal{E}$ using the associated $\mathcal{E}_p^t$ prototype subset.
\end{enumerate}
}}
\caption{The Evolutionary Kernel Machine: a co-evolution framework}
\label{fig:CoevolAlgorithm}
\end{center}
\end{figure}

More precisely, a co-evolutionary framework involving three species is considered, as detailed in Figure \ref{fig:CoevolAlgorithm}.
The first species includes the GP kernels. The second species includes the prototype subset (fixed-size subsets of the training set), subject to a cooperative co-evolution \cite{Potter2000} with the GP kernels. The third species includes the fitness case subset (fixed-size subsets of the training set), subject to a competitive host-parasite co-evolution \cite{Hillis1990} with the GP kernels.

The prototype species is evolved to find good prototypes such that they maximize the fitness of the GP kernels. The fitness case species is evolved to find hard and challenging examples, such that they minimize the kernel fitness. Of course there is a danger that the fitness case subset ultimately capture the noisy examples, as observed in the boosting framework \cite{Freund1996} (see Section \ref{discu}).

Both prototype and selection species are initialized using a stratified uniform sampling with no replacement (the class distribution in the sample is the same as in the whole dataset and all examples are distinct). Both species are evolved using a $(1,\lambda)$ evolution strategy; in each generation, $\lambda$ offsprings are generated using a uniform stratified replacement of a given fraction of the parent subset, and assessed after the best kernel in the current kernel population. The parent subset is replaced by the best offspring. In each generation, the kernels are assessed after the current prototype and fitness case individuals.

\section{Experimental Validation}
\label{sec:Results}

This section reports on the experimental validation of EKM, on a standard set of benchmark problems \cite{Newman1998}, detailed in Table \ref{tab:UCIDataSetDescription}. The system is implemented using the Open BEAGLE framework\footnote{\url{http://beagle.gel.ulaval.ca}} for evolutionary computation \cite{Gagne2006}.
\begin{table}[tb]
\caption{UCI data sets used for the experimentations.}
\label{tab:UCIDataSetDescription}
\begin{center}
\begin{tabularx}{\linewidth}{c|c|c|c|X}
Data      &        & \# of    & \# of   &\\
set       & Size   & features & classes & Application domain\\\hline
{\tt bcw} & $683$  & $9$      & $2$     & Wisconcin's breast cancer, $65\:\%$ benign and $35\:\%$ malignant.\\
{\tt bld} & $345$  & $6$      & $2$     & BUPA liver disorders, $58\:\%$ with disorders and $42\:\%$ without disorder.\\
{\tt bos} & $508$  & $13$     & $3$     & Boston housing, $34\:\%$ with median value $v<18.77$ K\$, $33\:\%$ with $v\in]18.77,23.74]$, and $33\:\%$ with $v>23.74$.\\
{\tt cmc} & $1473$ & $9$      & $3$     & Contraceptive method choice, $43\:\%$ not using contraception, $35\:\%$ using short-term contraception, and $23\:\%$ using long-term contraception.\\
{\tt ion} & $351$  & $34$     & $2$     & Ionosphere radar signal, $36\:\%$ without structure detected and $64\:\%$ with a structure detected.\\
{\tt pid} & $768$  & $8$      & $2$     & Pima indians diabetes, $65\:\%$ tested negative and $35\:\%$ tested positive for diabetes.\\
\end{tabularx}
\end{center}
\end{table}
\subsection{Experimental Setting}
The parameters used in EKM are reported in Table \ref{tab:Parameters}. The average evolution time for one run is less than one hour (AMD Athlon 2800+).

On each problem, EKM has been evaluated along the standard 10-fold cross validation methodology. The whole data set is partitioned into 10 (stratified) subsets; the training set is made of all subsets but one; the best hypothesis learned from this training set is evaluated on the remaining subset, or test set. The accuracy is averaged over the 10 folds (as the test set ranges over the 10 subsets of the whole dataset); for each fold, EKM is launched 10 times; the 5 best hypotheses (after their accuracy on the training set) are assessed on the test set; the reported accuracy is the average over the 10 folds of these 5 best hypotheses on the test set. In total, EKM is launched 100 times on each problem.
\begin{table}[tb]
\caption{Tableau of the evolutions parameters.}
\label{tab:Parameters}
\begin{center}
\begin{tabularx}{\linewidth}{c|X}
Parameter                & Description and parameter values\\\hline
\multicolumn{2}{c}{GP kernel functions evolution parameters}\\\hline
Primitives               & See Table \ref{tab:GPPrimitives}.\\
GP population size       & One population of $p=1000$ individuals\\
Stop criterion           & Evolution ends after $T=100$ generations.\\
Replacement strategy     & Genetic operations applied following generational scheme.\\
Selection                & Lexicographic parsimony pressure tournaments selection with $7$ participants.\\
Crossover                & Classical subtree crossover \cite{Koza1992} (prob. $0.7$).\\
Standard mutation        & Crossover with a random individual (prob. $0.1$).\\
Swap node mutation       & Exchange a primitive with another of the same arity (prob. $0.1$).\\
Shrink mutation          & Replace a branch with one of its children and remove the branch mutated and the other children subtrees (if any) (prob. $0.1$).\\\hline
\multicolumn{2}{c}{Prototype subset selection parameters}\\\hline
Prototype subset size    & $\ell=50$ examples in a prototype subset.\\
Number of offsprings     & $\lambda_p=4$ offsprings per generation.\\
Mutation rate            & $\rho_p=25\:\%$ of the prototype examples replaced in each mutation.\\\hline
\multicolumn{2}{c}{Fitness case subset selection parameters}\\\hline
Fitness case subset size & $m=100$ examples in a fitness case subset.\\
Number of offsprings     & $\lambda_s=2$ offsprings per generation.\\
Mutation rate            & $\rho_s=50\:\%$ of the selection examples replaced in each mutation.\\
\end{tabularx}
\end{center}
\end{table}

EKM is compared to state of the art algorithms, including $k$-nearest neighbor and SVMs with Gaussian kernels, similarly assessed using 10-fold cross validation. For $k$-NN, the underlying distance is the Euclidean one, and scaling normalization option has been considered; the $k$ parameter has been varied in $\{1,3,5\}$; the best setting has been kept. For Gaussian SVMs, the Torch3 implementation has been used \cite{Collobert2002}; the error cost (parameter $C$) has been varied in $\{10^{i},~i=-3\ldots 4\}$, the $\sigma$ parameter is set to $10$, and the best setting has been similarly retained. 

\subsection{Results}
\label{discu}

Table \ref{tab:UCIDataSetResults} shows the results obtained by EKM compared with $k$-NN and Gaussian SVM, together with the optimal parameters for the latter algorithms. 
\begin{table}[tb]
\caption[Results on the UCI data sets.]{Comparative 10-fold results of $k$-NN, Gaussian SVM and EKM on the UCI data sets, with optimal settings ($k$ and scaling for $k$-NN, $C$ for SVM). The reported test error is averaged over the 10 folds. For each fold tested with the EKM, the 5 solutions out of 10 runs with best training error are assessed on the test set, and their error is averaged. Test error rates in {\bf bold} denotes the statistically best results according to a 95\% two-tails paired Student's $t$-test. ``Average rank'' column gives the test error ranking obtained for EKM compared to $k$-NN and SVM averaged over the $10$ folds.} 
\label{tab:UCIDataSetResults}
\begin{center}
\begin{tabular}{c|c|c|c|c|c|c|c|c|c|c|c}
          & \multicolumn{4}{c|}{$k$-NN}                                  & \multicolumn{3}{c|}{SVM}               & \multicolumn{4}{c}{EKM}\\
Data      & \multicolumn{2}{c|}{Best conf.} & Train   & Test             & Best    & Train     & Test             & Train   & Best-half        & Mean & Average\\
set       & $k$ & Scaling                   & error   & error            & $C$     & error     & error            & error   & test error       & size & rank\\\hline
{\tt bcw} & $5$ & No                        & $0.027$ & $\mathbf{0.025}$ & $1$     & $0.030$   & $\mathbf{0.028}$ & $0.020$ & $\mathbf{0.030}$ & $167$ & $2.1$\\
{\tt bld} & $5$ & No                        & $0.336$ & $0.353$          & $1$     & $0.329$   & $\mathbf{0.325}$ & $0.299$ & $\mathbf{0.309}$ & $158$ & $1.5$\\
{\tt bos} & $1$ & Yes                       & $0.248$ & $\mathbf{0.235}$ & $0.001$ & $0.224$   & $0.308$          & $0.253$ & $0.281$          & $116$ & $1.8$\\
{\tt cmc} & $5$ & No                        & $0.491$ & $0.486$          & $10$    & $0.273$   & $\mathbf{0.433}$ & $0.479$ & $0.487$          & $129$ & $2.4$\\
{\tt ion} & $1$ & Yes                       & $0.134$ & $0.134$          & $100$   & $0.070$   & $\mathbf{0.071}$ & $0.078$ & $\mathbf{0.095}$ & $156$ & $1.9$\\
{\tt pid} & $5$ & Yes                       & $0.265$ & $\mathbf{0.255}$ & $0.001$ & $0.315$   & $0.307$          & $0.237$ & $\mathbf{0.252}$ & $145$ & $1.45$\\
\end{tabular}
\end{center}
\end{table}
The size of the best GP kernel (last column) shows that no bloat occurred, thanks to the lexicographic parsimony pressure. Each algorithm is shown to be the best performing on the half or more of the tested datasets, with frequent ties according to a paired Student's $t$-test.

Typically, the problems where Gaussian SVMs perform well are those where the optimal $C$ value for cost error is high, suggesting that the noise level in these datasets is high too. Indeed, the fitness case subset selection embedded in EKM might favor the selection of noisy examples, as those are more challenging to GP kernels. A more progressive selection mechanism, taking into account all kernels in the GP population to better filter out noisy examples and outliers, will be considered in further research.

The $k$-NN outperforms SVM and EKM on the {\tt bos} problem, where the noise level appears to be very low. Indeed, the optimal value for the number $k$ of nearest neighbors is $k=1$, while the optimal cost error is $10^{-3}$, suggesting that the error rate is also low. Still, the fact that the error rate is close to 23\% might be explained as the target concept is complex and/or many examples lie close to its frontier. On {\tt bcw}, the differences between the three algorithms are not statistically different and the test error rate is about 2\%, suggesting that the problem is rather easy.

EKM is found to outperform the other algorithms on {\tt bld}, demonstrating that Kernel-based dissimilarity can improve on Euclidean distance with and without rescaling. Last, EKM behaves like $k$-NN on the {\tt pid} problems. Further, it must be noted that EKM classifies the test examples using a 50-examples prototype set, whereas $k$-NN uses the whole training set (above 300 examples in the {\tt bld} problem and 690 in the {\tt pid} problem).

As the well-known No Free Lunch theorem applies to Machine Learning too, no learning method is expected to be universally competent. Rather, the above experimental validation demonstrates that the GP-evolved kernels can improve on standard kernels in some cases.

\section{Related Works}
\label{sec:RelatedWorks}

The most relevant work to EKM is the Genetic Kernel Support Vector Machine (GK-SVM) \cite{Howley2005}. GK-SVM similarly uses GP within an SVM-based approach, with two main differences compared to EKM. On one hand, GK-SVM focuses on feature construction, using GP to optimize mapping $\Phi$ (instead of the kernel). On the other hand, the fitness function used in GK-SVM suffers from a quadratic complexity in the number of training examples. Accordingly, all datasets but one considered in the experimentations are small (less than 200 examples). On a larger dataset, the authors acknowledge that their approach does not improve on a standard SVM with well chosen parameters. Another related work similarly uses GP for feature construction, in order to classify time series \cite{Eads2002}. The set of features (GP trees) is further evolved using a GA, where the fitness function is based on the accuracy of an SVM classifier. Most other works related to evolutionary optimization within SVMs (see \cite{Friedrichs2005}) actually focus on parametric optimization, e.g. achieving features selection or tuning some parameters.

Another related work is proposed by Weinberger {\em et al.} \cite{Weinberger2005}, optimizing a Mahalanobis distance based on the $k$-NN margin criterion inspired from \cite{Gilad-Bachrach2004} and also used in EKM. However, restricted to linear changes of representation, the optimization problem is tackled by semi-definite programming in \cite{Weinberger2005}. Lastly, EKM is also inspired by the Dynamic Subset Selection first proposed by Gathercole and Ross \cite{Gathercole1994} and further developed by \cite{Song2005} to address scalability issues in EC-based Machine Learning.

\section{Conclusion}
\label{sec:Conclusion}

The Evolutionary Kernel Machine proposed in this paper aims to improve kernel-based nearest neighbor classification \cite{Yu2002}, combining two original aspects. First, EKM implicitly addresses the feature construction problem by designing a new representation of the application domain better suited to the dataset at hand. However, in contrast with \cite{Howley2005,Eads2002}, EKM takes advantage of the kernel trick, using GP to optimize the kernel function. Secondly, EKM proposes a co-evolution framework to ensure the scalability of the approach and control the computational complexity of the fitness computation. The empirical validation demonstrates that this new approach is competitive with well-founded learning algorithms such as SVM and $k$-NN using tuned hyper-parameters. 

A limitation of the approach, also observed in the well-known boosting algorithm \cite{Freund1996}, is that the competitive co-evolution of kernels and examples tends to favor noisy validation examples. A perspective for further research is to exploit the evolution archive, to estimate the probability for an example to be noisy and achieve a sensitivity analysis. Another perspective is to incorporate ensemble learning, typically bagging and boosting, within EKM. Indeed the diversity of the solutions constructed along population-based optimization enables ensemble learning almost for free.

\subsection*{Acknowledgments}

This work was supported by postdoctoral fellowships from the ERCIM (Europe) and the FQRNT (Qu\'ebec) to C. Gagn\'e. M. Schoenauer and M. Sebag gratefully acknowledge support by the PASCAL Network of Excellence, IST2002506778.

\bibliographystyle{splncs}
\bibliography{evokern-paper}

\end{document}